\documentclass[a4paper, 10pt, conference]{ieeeconf}      
\usepackage{FG2024}

\FGfinalcopy 

\IEEEoverridecommandlockouts                              
\usepackage{graphicx}
\usepackage{amsmath}
\usepackage{amssymb}
\usepackage{booktabs}

\usepackage{cite}
\usepackage{color}
\usepackage{float}
\usepackage{caption}
\usepackage{tabularx}
\usepackage{array}
\newcolumntype{P}[1]{>{\centering\arraybackslash}p{#1}}
\usepackage{hhline}
\usepackage[ruled,linesnumbered]{algorithm2e}
\usepackage{dblfloatfix}
\usepackage{multirow}

\def\FGPaperID{****} 

\title{\LARGE \bf
SaFL: Sybil-aware Federated Learning with Application to Face Recognition
}

\author{Mahdi Ghafourian, Julian Fierrez, Ruben Vera-Rodriguez, Ruben Tolosana, Aythami Morales\\
Biometrics and Data Pattern Analytics, BiDA-Lab, Universidad Autonoma de Madrid, Spain\\
{\tt\small (mahdi.ghafourian,julian.fierrez,ruben.vera,ruben.tolosana,aythami.morales)@uam.es}}

\begin{document}

\ifFGfinal
\thispagestyle{empty}
\pagestyle{empty}
\else
\author{Anonymous FG2024 submission\\ Paper ID \FGPaperID \\}
\pagestyle{plain}
\fi
\maketitle

\begin{abstract}

Federated Learning (FL) is a machine learning paradigm to conduct collaborative learning among clients on a joint model. The primary goal is to share clients' local training parameters with an integrating server while preserving their privacy. This method permits to exploit the potential of massive mobile users' data for the benefit of machine learning models' performance while keeping sensitive data on local devices. On the downside, FL raises security and privacy concerns that have just started to be studied. To address some of the key threats in FL, researchers have proposed to use secure aggregation methods (e.g. homomorphic encryption, secure multiparty computation, etc.). These solutions improve some security and privacy metrics, but at the same time bring about other serious threats such as poisoning attacks, backdoor attacks, and free running attacks. This paper proposes a new defense method against poisoning attacks in FL called SaFL (Sybil-aware Federated Learning) that minimizes the effect of sybils with a novel time-variant aggregation scheme.

\end{abstract}

\section{INTRODUCTION}
Traditionally, the performance of machine learning models depends on the computational power and the availability of training data in a centralized server. However, optimizing both elements (computing power and amount of data) in a centralized server is quite challenging in most applications, among other technical reasons, due to security \cite{9064510,2023_Book-PAD} and privacy requirements \cite{2022_Access_DP-CL_Ahmad,2017_Access_HEmultiDTW_Marta}. Particularly, the demands for high-quality machine learning models on one hand, and tensions for user privacy on the other hand are driving research and technology to propose methods to improve the performance of models without exposing sensitive information \cite{2021_TPAMI_SensitiveNets_Morales,pena2023human}. In 2017, McMahan \textit{et al.} \cite{mcmahan2017communication} proposed an approach called Federated Learning (FL) in which regular people can use their local data stored on their mobile devices for training a shared machine learning model in a distributed way by aggregating locally-computed gradient updates \cite{zhang2021survey}. The main insight behind FL is to guarantee the privacy of participants by sharing local training model parameters instead of their actual data. However, there are multiple security and privacy concerns threatening FL \cite{lyu2020threats, jere2020taxonomy}. Specifically, recent studies have reported different innate vulnerabilities in the FL framework (e.g., gradient leakage \cite{zhu2020deep, wei2020framework,mothukuri2021survey}) which can be exploited by adversaries to compromise participant's data privacy or system robustness.

\begin{figure}[t]
\begin{center} \includegraphics[trim={10cm 4.5cm 3cm 3cm},clip,width=1.4\linewidth]{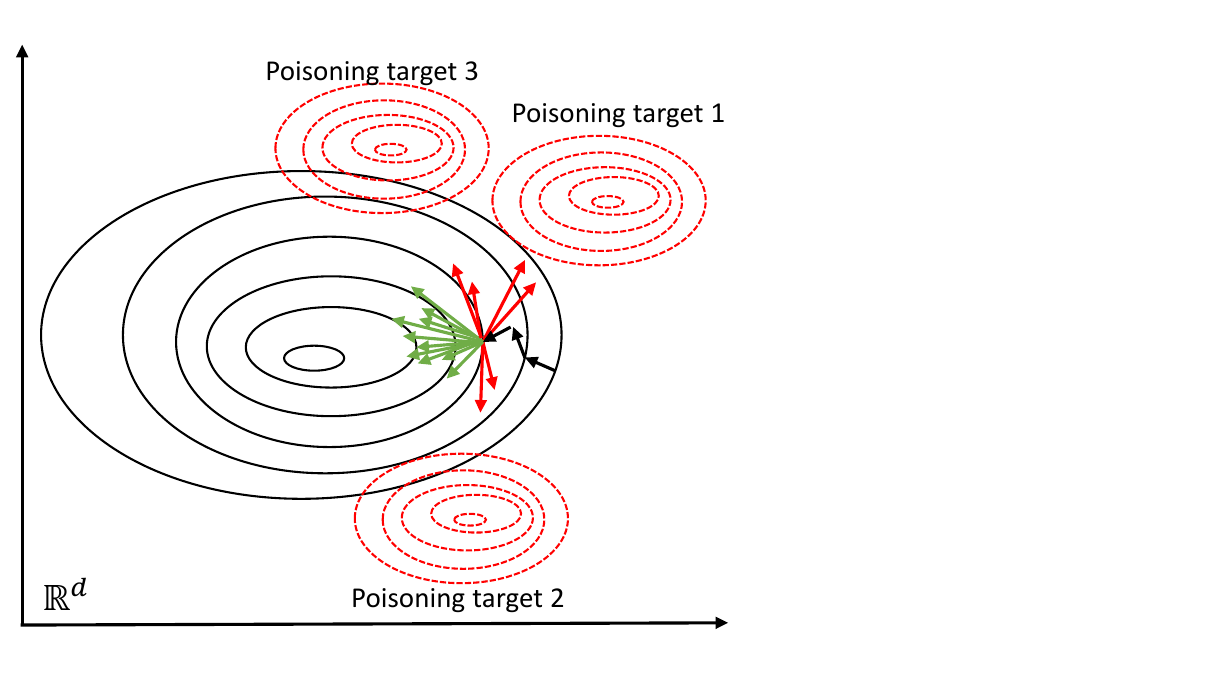}
\end{center}
   \caption{Visual concept of a targeted poisoning attack with multiple targets in Federated Learning (FL). The big oval represents the true objective of the aggregation. Black arrows show the first learning steps in the $\mathbb{R}^{d}$ space of learning parameters. The red arrows are sybils with different targets. Green arrows are updates from benign clients.}
\label{fig:Fig1}
\end{figure}

To address these threats, researchers have proposed to use cryptography-based techniques such as \textit{Homomorphic Encryption} \cite{lyu2018ppfa}, \textit{Secure Multiparty Encryption} (SMC), \textit{Secret Sharing}, and other secure aggregation methods to deliver the average gradients of all participants to the aggregator. However, most of the proposed cryptographic techniques open the door to other malicious activities and threats. Besides, they are useless against adversarial attacks \cite{ghafourian2023toward} in deep learning. Poisoning attacks, backdoor attacks, and free-riding are the most important challenges that have been reported so far \cite{jere2020taxonomy, lyu2022privacy} for FL. Particularly, targeted poisoning enables an adversary to poison a specific class in the training process so as to increase or decrease the probability of the trained model predicting a sample of the target class. For example, in terms of face recognition \cite{ghafourian2022otb}, this attack helps the adversary to impersonate or avoid fraud detection \cite{2023_Book-PAD_Face,2022_Handbook_IntroFaceManipulation_RT}.
\begin{figure*}[b!]
 \centering 
 \includegraphics[trim={0.9cm 5cm 0.7cm 1.5cm},clip,width=170mm,scale=0.5]{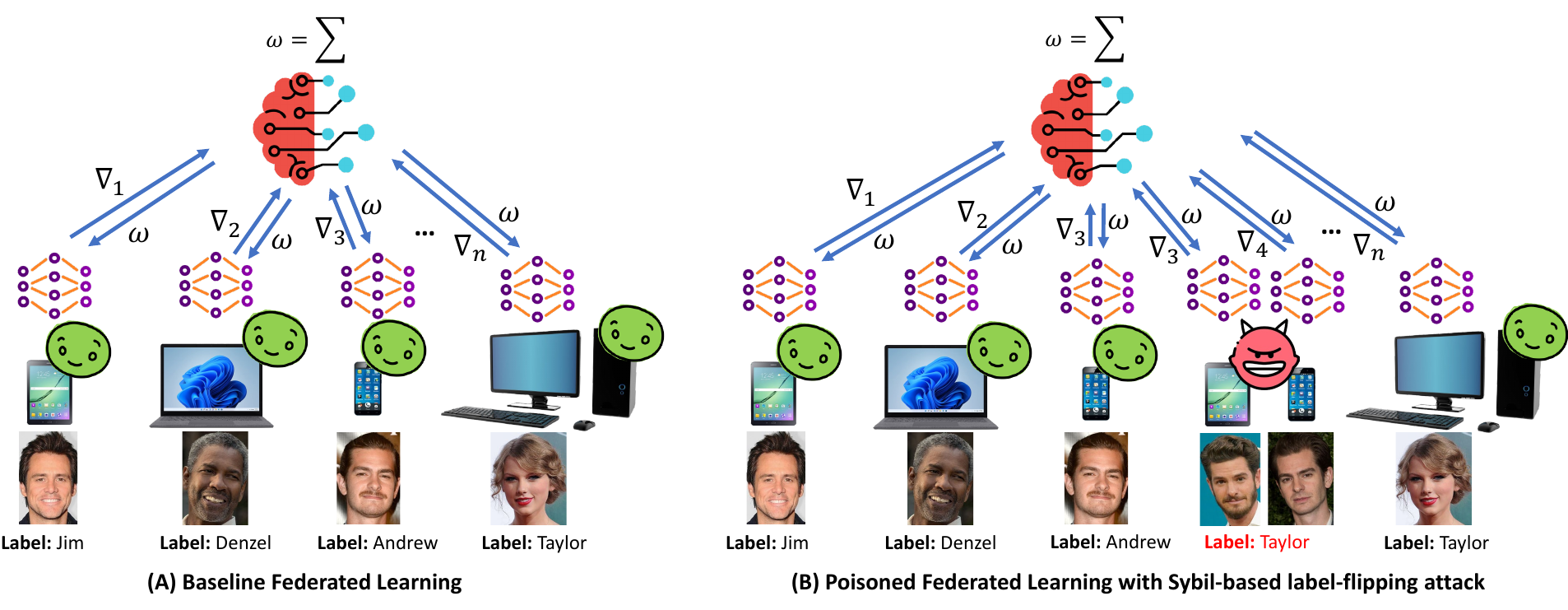}
  \caption{Federated Learning paradigm for face recognition using non-IID data with and without Sybil-based poisonous attack. The label-flipping attack is demonstrated in (b) where two Sybils poison the model by training locally on their arbitrary data but labeled with the victim (target).}
  \label{fig:Fig2}
\end{figure*}

Recent attempts to address this threat have dealt with it as the classic Byzantine generals' problem \cite{lamport2019byzantine} or Sybil attack, where traitor Byzantine generals enter the system with multiple agents. These agents have colluded to overpower the model to their malicious target. In addition, it is also possible that multiple Byzantines who have different malicious targets from other adversaries exist in the system. Therefore, any solution to overcome this threat in federated learning must address both single target (same target) and multi-target (different target) poisoning attacks. Fig. \ref{fig:Fig1} depicts the federated learning paradigm where multiple sybils with distinguished poisoning targets are involved in the training process.

This paper focuses on addressing targeted poisoning attacks in a general federated learning framework with application to face recognition where clients train a global model locally and send their training parameters to a central server to be aggregated and hence update the global model. In particular, the contributions of this paper are as follows:
\begin{itemize}
\item We propose a new defense method against poisoning attacks in FL that minimizes the impact of sybils by combining in a novel way two main ideas: 1) not considering client updates out of a certain range and 2) varying that range in time to take into account the progressive learning that typically happens in FL architectures.
\item The effect of the proposed approach is compared with two popular defense methods in terms of protection rate and learning performance.
\end{itemize}


\section{Background and related works}
\label{sec:relatedworks}
Federated Learning (FL) is a collaborative learning paradigm that enables a multitude of participants to train a ML model jointly without exposing their private data (Fig. \ref{fig:Fig2} (A)). 
The resulting trained model can be poisoned by different approaches (in our case label flipping attack). It is possible that some adversaries attend the joint training with multiple agents (sybils) and send their malicious updates to be aggregated with those of benign participants so as to overpower the training process to poison a target class. To mitigate this, two noteworthy methods (Multi-Krum\cite{blanchard2017machine}, and FoolsGold\cite{fung2020limitations}) have been proposed recently to exclude sybils from the aggregation. In this section, we give a brief overview of these two important methods and discuss their limitation.

\subsection{Multi-Krum}

This research argues that no linear aggregation (e.g., Fed-avg \cite{mcmahan2017communication}) can tolerate a single Byzantine as they can prevent a simple averaging-based approach to converge by forcing the aggregating server to choose their arbitrary vector regardless of its amplitude and distance from other vectors (updates). To address this problem, they propose an aggregation method that calculates the Euclidean norm on the $N-f-2$ closest vectors to each vector $V_{i}$, where $N$ is the total number of participants of FL and $f$ is the number of Byzantines. Eventually, it selects one vector that minimizes the Euclidean norm to its $N-f-2$ closest vectors as the result of the aggregation.
However, this method does not work well on federated learning since when we start training a model from scratch, it hasn't learned anything yet and thus the gradients received from participants are very inaccurate and far from each other. Therefore, \textit{MK} mistakenly excludes more than necessary gradients of honest participants from the aggregation, and thus the resulting model won't learn the corresponding classes for the excluded gradients.

\subsection{FoolsGold}
Fung et al. \cite{fung2020limitations} proposed an aggregation method called \textit{FoolsGold (FG)} that improves some of the weaknesses of \textit{MK}. This method finds a learning rate for each received gradient (accumulated gradients) to be multiplied by when aggregating. The learning rate is the smallest for those vectors that are nearest to each other based on their cosine distance. To prevent the key limitations of \emph{MK} (i.e., discarding benign gradients), they included a pardoning term that avoids penalizing honest participants. However,  \textit{FG} only works when the number of Sybils $> 1$. Additionally, an attacker can simply avoid the exclusion of his Sybils' gradients by sending two dissimilar updates. One update is similar (close in terms of cosine distance) to that of the victim (true class) obtained by eavesdropping from the communication channel and the other update contains the poisonous gradient. Since \textit{FG} excludes updates that are closest to each other, it excludes the victim's update and that of the attacker which is similar. Then, it updates the target class with the remaining poisonous update. This threat and how the proposed mitigation method addresses it is depicted in Fig. \ref{fig:Fig3} (B).


\section{Threat model and assumptions}
\label{sec:threatmodel}
The goal of the past works was to exclude poisonous updates from the aggregation (i.e., poisoning prevention). However, this goal cannot be achieved completely in FL.

In general, there are two scenarios to poison federated learning:
\begin{enumerate}
\item Sybils join the federated learning from scratch
\item Sybils join the training later after some iterations
\end{enumerate}

The difficulty with finding the Sybils in the former scenario is that the model already hasn’t learned anything about each class when the training starts. Therefore, the gradients of even benign users are quite divergent. In previous works, \textit{FG} and \textit{MK} applied their protection method from the first iterations. In federated learning settings where each class of the model represents the data of one client, an impostor who joins the training with multiple Sybils can overpower the target class to his poisonous target. 
In fact, at least in non-IID settings, using \textit{MK} can accelerate learning poisonous data by excluding gradients of the benign clients. \textit{FG} also cannot mitigate this problem, as we discussed earlier.

To poison the target class in our experiments, the malicious client does the label-flipping attack (see Fig. \ref{fig:Fig2} (B)). To this end, he has a dataset of his face labeled as the victim's class (target class). The successful attack generates a model that incorrectly classifies the attacker as the victim. The attack success rate that we compute in our experiments is the proportion of the attacker faces from the test set being mistakenly classified as the victim by the final model. 

In this paper, the attacker capabilities and general assumptions are as follows:

\begin{itemize}
\item The attacker is able to join and leave FL with multiple colluding agents (Sybils) whenever he wants.
\item The attacker is able to obtain the updates of some honest participants by eavesdropping or other malicious means.
\item The attacker cannot exploit the gradient-leakage attack to reconstruct the training data of honest participants.
\item The attacker is able to conduct targeted poisoning by performing the label-flipping attack.
\item The aggregator server is an honest but curious uncompromised party.
\item At least one benign update per class exists in each aggregation iteration.
\item FL participants are permitted to send multiple updates from training their data on multiple devices they have.
\end{itemize}


\section{Sybil-aware Federate Learning (SaFL)}
\label{sec:proposedmethod}
\subsection{SaFL aggregation scheme}
Any solution to minimize the adverse effect of the poisoning attack or backdoor attack in the Federated Learning framework should satisfy the following goals:

\begin{enumerate}
\item The mitigation method should not lower the performance of the joint model nor prevent learning benign updates regardless of the existence of poisoning updates.
\item The mitigation method should be robust to the number of Sybils.
\item The mitigation method should be robust to multiple poisoning targets.
\end{enumerate}

\begin{figure}[b!]
\begin{center} \includegraphics[trim={13cm 4.8cm 9.2cm 4.2cm},clip,width=1.0\linewidth]{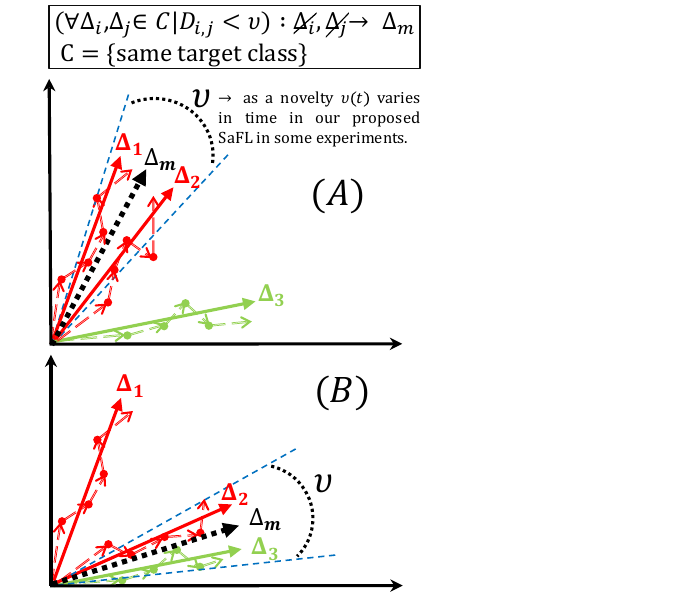}
\end{center}
   \caption{Solid arrows represent accumulated updates of participants (red arrows show Sybils, green shows benign update). Black dashed line shows the median of all accumulated updates within $\nu$ at iteration $t$. Dashed arrows around solid lines are the gradients of previous iterations until the iteration $t$. Our SaFL method mitigates sybil attack in scenario (A) where the attacker uses two sybils with very similar updates to overpower the targeted class and in (B) where the attacker tries to fool the mitigation method so as to cause the deletion of benign update(s) from aggregation.}
\label{fig:Fig3}
\end{figure}

The main limitation of \textit{MK} and \textit{FG} is that both these methods are based on forming clusters on participants' updates excluding those that fall out of the main cluster. However, due to the dramatic change of gradients in initial training epochs, considering only a main gradient cluster may provoke the removal of significant honest data. The pardoning mechanism in \textit{FG} to overcome this issue is not helpful since the attacker can simply avoid detection as we discussed earlier. 

\RestyleAlgo{ruled}
\SetKwComment{Comment}{/* }{ */}

\begin{algorithm}[t!]
\caption{The SaFL aggregation method}\label{alg:one}
\KwData{Initial Model $w_{0}$ and SGD updates ${\nabla}_{i,t}$ from each client $i$ at iteration $t$.}
\For{iteration $t=1 \dots T$}{
  \For{all clients $i$}{           ${\Delta}_{i}=\sum_{\tau=1}^{t} {\nabla}_{i,\tau}$\\
  }
  \For{all clients $i$}{ 
        \For{all clients $j$}{ 
            ${D}_{i,j}=1-\frac{{\Delta}_{i}\times{\Delta}_{j}}{\lVert {\Delta}_{i} \rVert \times \lVert {\Delta}_{j} \rVert}$\\
             \If{${D}_{i,j} < {\nu}$ \textbf{and} $i,j \notin GL$}{
             Generate a new group $G$\\
             Add ${\Delta}_{i}, {\Delta}_{j}$ to $G$\\
             Add $G$ to the group list $GL$\\
             }
        }
  }
  \For{all clients $i$}{           
    \If{${\Delta}_{i} \notin GL$}{
            Add ${\Delta}_{i}$ to ${\gamma}_{t}$\\
          }
  }
  \For{each group $G \in GL$}{  
      \For{all elements $k \in G$}{           
            $m = \textrm{med}({\Delta}_{k})$\\
            Add $m$ to ${\gamma}_{t}$
      }
  }  $w_{t}=\textrm{UpdateModel}(w_{t-1},\gamma_t)$\\
\textbf{Output:} global model $w_{t}$
}
\end{algorithm}

Despite our argument, it is possible to minimize the effect of the poisoning attack by decreasing the influence of Sybils. The main idea behind our proposed aggregation scheme to overcome this limitation is to keep all the updates for the aggregation except those that are in a certain cosine distance from one another (i.e., those vectors $V_{i}$ and $V_{j}$ that their pair-wised cosine distance is below a specific degree of freedom $\nu$). For these vectors, the median will be selected and added to the aggregation. 

Our proposed SaFL aggregation ensures that all classes in the join model learn uniformly as it always takes one update as the representative of its cosine-adjacent vectors. Fig. \ref{fig:Fig3} shows how the SaFL method mitigates targeted poisoning under two scenarios where two Sybils targeting the same objective try to overpower one benign update. 

Let $w$ be the global aggregated model (joint model) in a FL with total iteration number $T$, ${\nabla}_{i,t}$ where $i \in\{1,...,n\}$ be the stochastic gradient descent (SGD) updates from each client $i$ at iteration $t$, ${\Delta}_{i}$ be the accumulated gradients of the client $i$ until time $t$, ${\nu}$ be the threshold (degree of freedom) of the pair-wise cosine distance $D_{i,j}$ between all received updates, $G$ be the group for aggregated updates of the same class where $D_{i,j} \leq \nu$, $GL$ be the list of groups $G$, $\textrm{med}$ be the element-wised median, and ${\gamma}_{t}$ be an array of updates that are being selected at iteration $t$. Then, our proposed SaFL aggregation method works as described in the Algorithm \ref{alg:one}.


\begin{figure*}[!t]
 \centering 
 \includegraphics[trim={2cm 4.5cm 2cm 4.5cm},clip,width=170mm,scale=0.5]{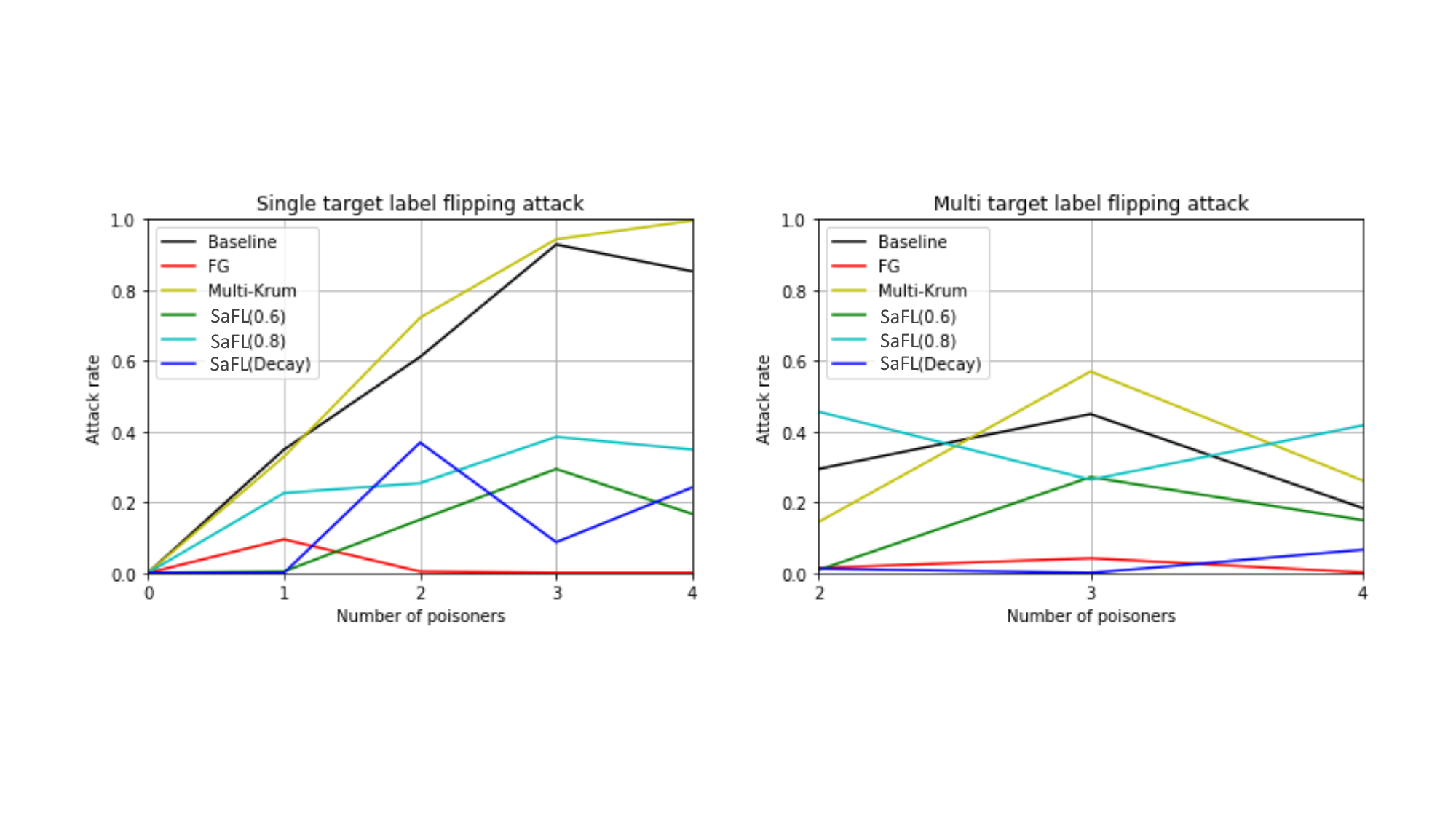}
  \caption{Label-flipping attack rate for a varying number of poisoners (Sybils) with a single target, left figure, and multiple targets (one Sybil per target), right figure using FL baseline, FoolsGold (FG), Multi-Krum, and SaFL at different thresholds.}
  \label{fig:Fig4}
\end{figure*}

\section{Evaluation}
\label{sec:evaluation}

For a fair evaluation and comparison, we implemented \emph{SaFL} and \emph{MK} on the \emph{FG} framework for federated learning\footnote{https://github.com/DistributedML/FoolsGold}. \textit{FG} uses 10 arbitrarily selected classes from the VGGFace2 \cite{21cao2018vggface2} dataset for the experiment on FL. We customized this framework for our experiments with the same settings. This is a collaborative training using squeeze-net \cite{iandola2016squeezenet} with non-IID data for 300 iterations where each participant only provides an update after locally training the global model on their corresponding class. Regarding malicious participants, they use the victim's data, label-flipped to the poisoning target for their Sybils. Unlike \textit{FG} which uses exactly the same poisoning data for all Sybils with the same target, in our experiments, we divided the source data between Sybils so it would be more realistic considering the intelligence of the attacker. Therefore, we compare the effectiveness of all methods using divided poisoning data. 
In general, our evaluation involves three experiments:

\begin{enumerate}
\item Comparing the attack success rate for mitigation methods plus the baseline considering two conditions of Sybils with single target and multiple targets.
\item Comparing the performance of all these methods regarding the collaborative training loss on the aforementioned conditions.
\item Computing the accuracy of our proposed estimated poisoning rate compared to the true poisoning rate.
\end{enumerate}

In all of our experiments, we evaluated our proposed \textit{SaFL} method considering three empirically selected thresholds $\nu\in\{0.6, 0.8, Decay\}$. Since in FL, the model starts learning from scratch, gradients of the same class are highly uncorrelated. Therefore, it's not possible to compute a similarity threshold to cluster correlated gradients. Considering the value of cosine distance ranging between 0 and 1, where the former indicates that two gradients are completely dissimilar and the latter denotes they are identical, we selected two initial heuristic thresholds of 0.8 as the strict threshold and 0.6 as the lenient threshold for our experiments. A third threshold \textit{Decay} is also computed by an exponential decay formula in each iteration as follows:
\begin{equation} \label{eq3}
    Decay(t) = \lambda(1-r)^{t}
\end{equation}
where $\lambda$ is the initial amount ($\lambda=0.8$ in our experiments to start with the strict threshold), $1-r$ is a decay factor ($r=0.001$ in our experiments), and $t$ is the iteration number.

The intuition behind the \textit{Decay} threshold is that as the model learns gradually, we decay the threshold so as to not mistakenly penalize the model and stop the learning to an extent where the model underfits. This might be important since in some first iterations taking the median of honest updates that their pairwise cosine distance is $<\nu$ won't affect the learning very much since the model already hasn't learned anything. However, as the learning proceeds, losing details by mistakenly discarding similar updates from honest clients restricts the learning.

\begin{figure}[b!]
\begin{center} \includegraphics[trim={4.6cm 3.2cm 1.6cm 2.7cm},clip,width=1.4\linewidth]{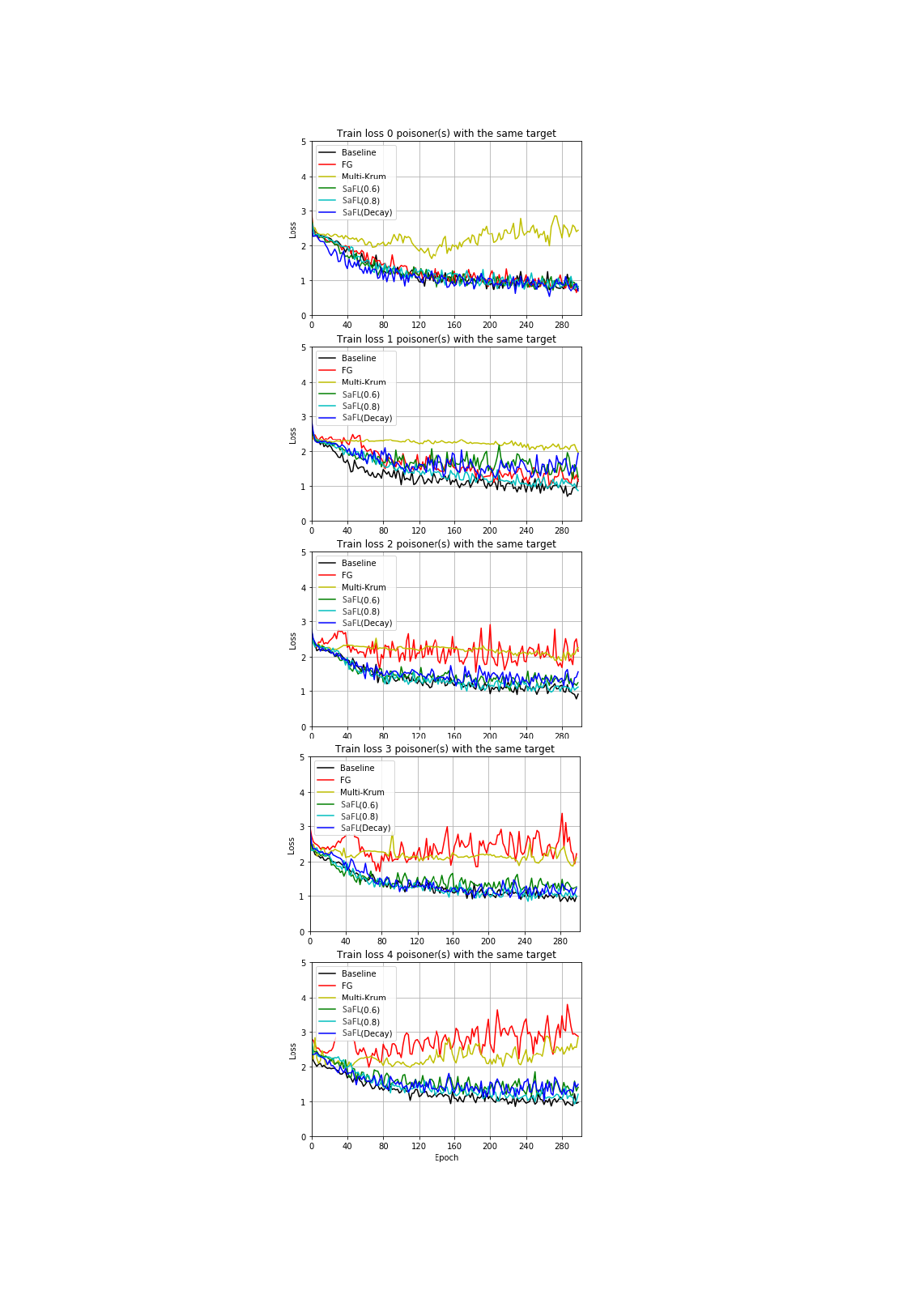}
\end{center}
  \caption{Training loss comparison over 300 iterations of federated learning considering the varying number of poisoners (Sybils) increasing from top to down with a single target.}
  \label{fig:Fig5}
\end{figure}
\begin{figure}[h!]
\begin{center} \includegraphics[trim={6cm 6.7cm 3cm 6.5cm},clip,width=1.4\linewidth]{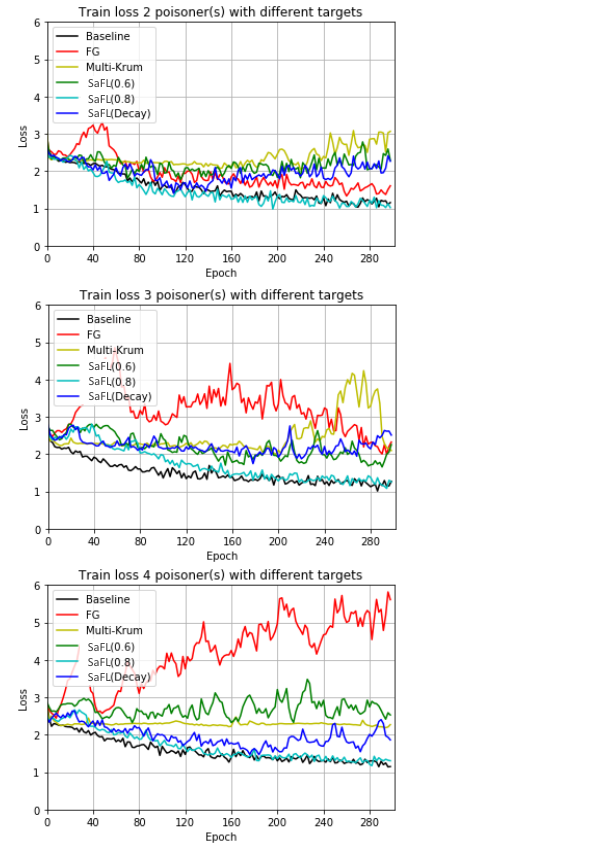}
\end{center}
  \caption{Training loss comparison over 300 iterations of federated learning considering varying numbers of poisoners (Sybils) increasing from top to down with multiple targets.}
  \label{fig:Fig6}
\end{figure}

\subsection{Attack success rate comparison}

In this experiment, 10 honest participants use their face images to train a \textit{SqueezeNet} model that has 10 classes representing the identity of each participant. Malicious participants can join this collaborative training with multiple Sybils and perform targeted poisoning. To this end, we conducted the label-flipping attack against unprotected FL and all aforementioned aggregation methods plus our proposed \textit{SaFL} method with different thresholds. The attack rate demonstrated in Fig. \ref{fig:Fig4} demonstrates the number of victims' faces from the test set which is classified as the poisoned label for Sybils with single target and multiple targets.

Regarding the single target, \textit{MK} with attack rate = 0.9 and 1.0 in the presence of 3 and 4 Sybils respectively, performs as poor as the unprotected FL (Baseline). Under the specific attack condition where Sybils' updates are in close distance of each other, \textit{FG} performs the best in both single-target and multi-target settings by lowering the attack rate to 0.0 regardless of the number of prisoners. In terms of our proposed \textit{SaFL}, while the result for the \textit{Decay} threshold oscillates slightly with different number of Sybils, \textit{SaFL} for $\nu=0.6$ performed better than other thresholds with the highest attack rate = 0.3 with 3 single targeted Sybils.

Except \textit{MK} which still reported the worse with attack rate $\simeq$ 0.6 for 3 Sybils aiming different targets, the result for Multi-target label flipping was unpredicted. While \textit{SaFL} for $\nu=0.6$ reported highest attack rate $\simeq$ 0.3 with 3 poisoners, \textit{SaFL} for $\nu=Decay$ almost performed as good as \textit{FG}.

\subsection{Performance comparison}
Referring to the first goal in Sec. \ref{sec:proposedmethod}, any targeted poisoning attack mitigation method is useless without maintaining the performance of the model. To this end, our second experiment aims at comparing all these aggregation methods in terms of their effect on the learning process of the model. Figs. \ref{fig:Fig5} and \ref{fig:Fig6} show our results for comparing training losses in 300 iterations of FL using the aforementioned aggregation methods with varying number of single-target and multi-target Sybils respectively. Looking at these figures, the general impression to notice is that even without any Sybils in the system, the baseline FL cannot reduce the training loss (Fig. \ref{fig:Fig5} first diagram from top) lower than 0.8 while fluctuating around 1.0 for almost 200 iterations. We also notice this trend for validation loss.

Looking at the compared aggregation methods, it is evident from graphs in Fig. \ref{fig:Fig5} that with the number of Sybils increasing, \textit{FG} performance decreases drastically to report the worst training loss $\simeq$ 3.8 when the number of Sybils is 4. On the other hand, \textit{MK} restricts learning with training loss $>$ 2.

Taking into account the proposed \textit{SaFL} method for single-targeted poisoning, when threshold $\nu=0.8$, the learning behavior in terms of training loss is very close to the Baseline (Similar behavior was also observed for validation loss, not reported here). This is because the $\nu=0.8$ is the less strict threshold among other experimented ones. The method is more easy-going with similar updates and thus it performed better than \textit{SaFL} with other experimented thresholds.

Looking at the graphs in Fig. \ref{fig:Fig6}, it is noticeable that \textit{FG} performs the worst with 4 multi-target Sybils as the training loss surges to loss $\simeq$ 6. In other words, \emph{FG} in a multi-targeted poisoning setup generates underfitting with the learning data. On the other hand, our proposed \emph{SaFL} is the method that best approximates the learning behavior in terms of training and validation loss under multi-targeted poisoning with respect to no attack at all.


\section{Conclusion}
In this work, we studied the effect of Sybil-based targeted poisoning via the label-flipping attack in FL. We particularly proposed a poisoning mitigation method and we compared it with the baseline (no protection) and two popular mitigation approaches. Our results show a high protection rate with negligible effect on the performance of the model, improving FL protection to withstand targeted poisoning attacks. Our results also confirm the efficacy of a decaying threshold to provide higher protection. However, the impact of different decay factors on both the protection rate and the model performance jointly should be studied further. 

For future work, we aim to investigate the potential benefits of meta-learning to tune our parameters (e.g., the decaying threshold parameters) in heterogeneous federated learning settings \cite{guo2022auto}. Finally, taking into account the heavy computational cost of current federated learning paradigms in real-world applications, it would be also fruitful to explore resource-adaptive federated learning \cite{NEURIPS2022_1b61ad02}.


\section{ACKNOWLEDGMENTS}
This work has been supported by projects: PRIMA (ITN-2019-860315), TRESPASS-ETN (ITN-2019-860813) and C\'atedra ENIA UAM-VERIDAS en IA Responsable (NextGenerationEU PRTR TSI-100927-2023-2).


{\small
\bibliographystyle{ieee}
\bibliography{egbib}
}

\end{document}